\RequirePackage[T1]{fontenc}
\documentclass[letterpaper,10pt,journal,twoside]{IEEEtran}
\usepackage{mathptmx,microtype}

\usepackage{color, soul}
\usepackage{amsmath,amssymb,bm,graphicx,booktabs,array}
\usepackage[table]{xcolor}
\usepackage{placeins,multirow}
\usepackage{tikz}
\usetikzlibrary{arrows.meta,positioning,calc}
\usepackage{cite}
\usepackage{xurl}
\usepackage[hidelinks]{hyperref}
\hypersetup{pdftitle={AHEADance: Anticipatory Humanoid Dance from Streaming Audio},pdfauthor={}}
\hypersetup{pdfauthor={Tianhu Peng, Yukun Wang, Lingfan Bao, Yiyang Jia, Chengxu Zhou}}
\definecolor{music}{HTML}{007F7A}
\definecolor{structure}{HTML}{315DA8}
\definecolor{detail}{HTML}{B76524}
\definecolor{state}{HTML}{6D568C}
\definecolor{pending}{HTML}{8B5921}
\newcommand{\method}{AHEADance}
\newcommand{\cof}{Commit Forcing}


\newcommand{\papertablestyle}{\centering\footnotesize\setlength{\tabcolsep}{3pt}\renewcommand{\arraystretch}{1.1}}
\newcommand{\tablenote}[1]{\par\vspace{3pt}\parbox{\linewidth}{\raggedright\scriptsize #1}}
\newcommand{\resultci}[3]{\ensuremath{#1\ [#2,\,#3]}}
\title{AHEADance: Anticipatory Humanoid Dance\\from Streaming Audio}
\author{Tianhu~Peng$^{*}$, Yukun~Wang$^{*}$, Lingfan~Bao, Yiyang~Jia and Chengxu~Zhou$^{\dagger}$\\[3pt]
\href{https://ihcr.github.io/AHEADance/}{\texttt{https://ihcr.github.io/AHEADance}}%
\thanks{This work was partially supported by the Advanced Research and Invention Agency (grant number SMRB-SE01-P06), the Isambard-AI National AI Research Resource \cite{isambard2025} and the NVIDIA Academic Grant Program using RTX PRO 6000.}%
\thanks{The authors are with the Department of Computer Science, University College London, London, U.K.}%
\thanks{$^{*}$Equal contribution. }
\thanks{$^{\dagger}$Corresponding author. \tt\small
chengxu.zhou@ucl.ac.uk}%
}
\IEEEoverridecommandlockouts
\makeatletter
\IEEEaftertitletext{%
  \vspace{-\baselineskip}%
  \begin{minipage}{\textwidth}
    \centering
    \setlength{\abovecaptionskip}{3pt}%
    \includegraphics[width=\textwidth]{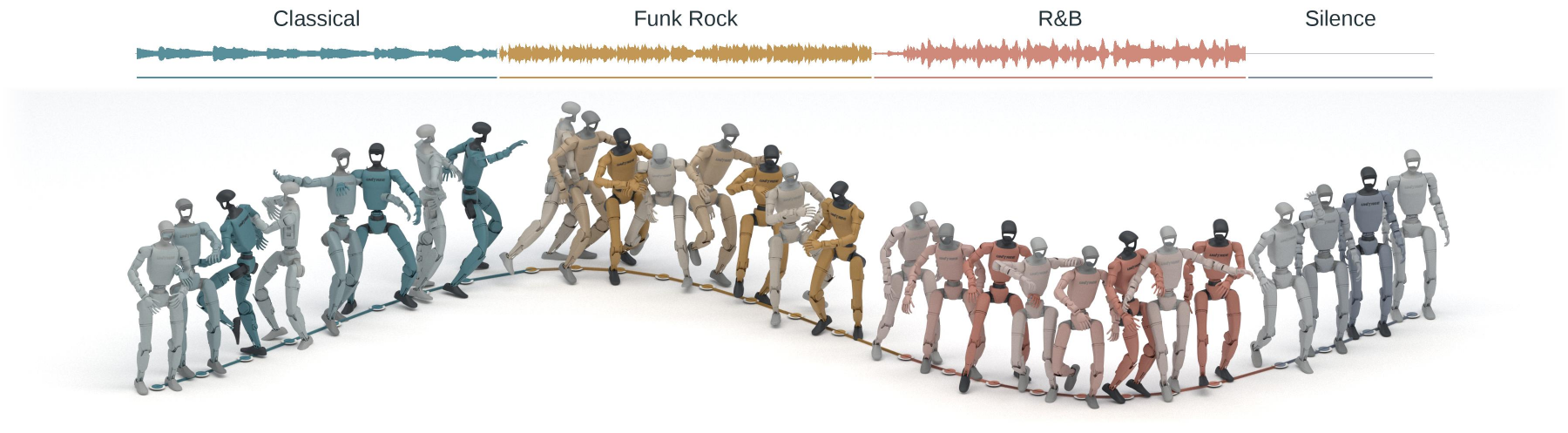}
    \def\@captype{figure}
    \caption{\textbf{Music-driven dance across styles and silence.}}
    \label{fig:streaming_teaser}
  \end{minipage}\vspace{0.3\baselineskip}%
}
\makeatother

\begin{document}
\bstctlcite{AHEADance:BSTcontrol}
\maketitle
\begin{abstract}
A humanoid that dances to streamed music must commit each motion chunk before
it has heard the music that chunk accompanies, and must continue from motion it
has already committed. We present \method{}, a robot-native streaming dance
generator executed on a humanoid robot by a whole-body tracker.
ListenAhead conditions each chunk on heard music and a frozen music
model's forecasts of the coming music, aligned to their target times and
weighted by reliability. The forecasts carry the timing of upcoming beats
almost as well as the actual future; with them, the generator is better
synchronized with the music and concentrates more motion energy at its song's
beat frequency. Commit Forcing trains continuation from the model's own
committed chunks, updating both the latent history and the decoded boundary of
the next chunk; it reduces low activity from 83.5\% to 10.3\%, the level of real
dance. Three music evaluators, each validated on real G1 dance, test
synchronization, song and style. On a physical Unitree G1, it dances while the music plays and comes to rest when it stops, without a fall.
\end{abstract}

\section{Introduction}
\label{sec:introduction}

A humanoid that dances to streamed music works under constraints that
offline choreography does not face. It hears only the music that has
already played, yet each motion chunk is generated from the music heard up to
the moment its own music begins, so that the delay from music to motion does not
grow with the chunk length. And committed
motion cannot be revised, so every new chunk has to continue from motion
the model itself produced rather than from a clean reference. Unlike
animation, every committed chunk is then executed by a whole-body tracking
controller on the robot, so it must also be trackable, and generation must
keep pace with the controller's clock. Under these constraints the robot must
keep dancing for as long as the music plays, come to rest when it stops and
dance again when it returns.

Music-to-dance generators, including long-form
ones~\cite{li2021aist,tseng2023edge,li2023finedance,li2025soulnet,li2024lodge},
receive the complete music in advance. Streaming humanoid systems hear the
music causally: DiscoForcing~\cite{ji2026discoforcing} streams with diffusion
forcing and retargets the generated human motion,
RoboPerform~\cite{li2026roboperform} maps audio to a control policy, and
InfiniDanceRobot~\cite{zhang2026infinidancerobot} closes the loop between a
planner and its tracker; none of them describes forecasting the coming music.
Autoregressive generators trained on reference histories meet their own outputs
at inference; diffusion forcing~\cite{chen2024diffusionforcing} and self
forcing~\cite{huang2025selfforcing} address this for the history the generator
conditions on, whereas a chunked robot generator also conditions on the decoded
pose boundary it has committed.
For a robot-native streaming generator, three questions remain:
how much of the coming music a causal generator can use, how to continue
reliably from its own committed chunks when both the latent history and
the decoded pose boundary change, and how to show that the executed
behavior follows the music rather than its own momentum.

\begin{figure*}[t]
\centering
\includegraphics[width=\textwidth]{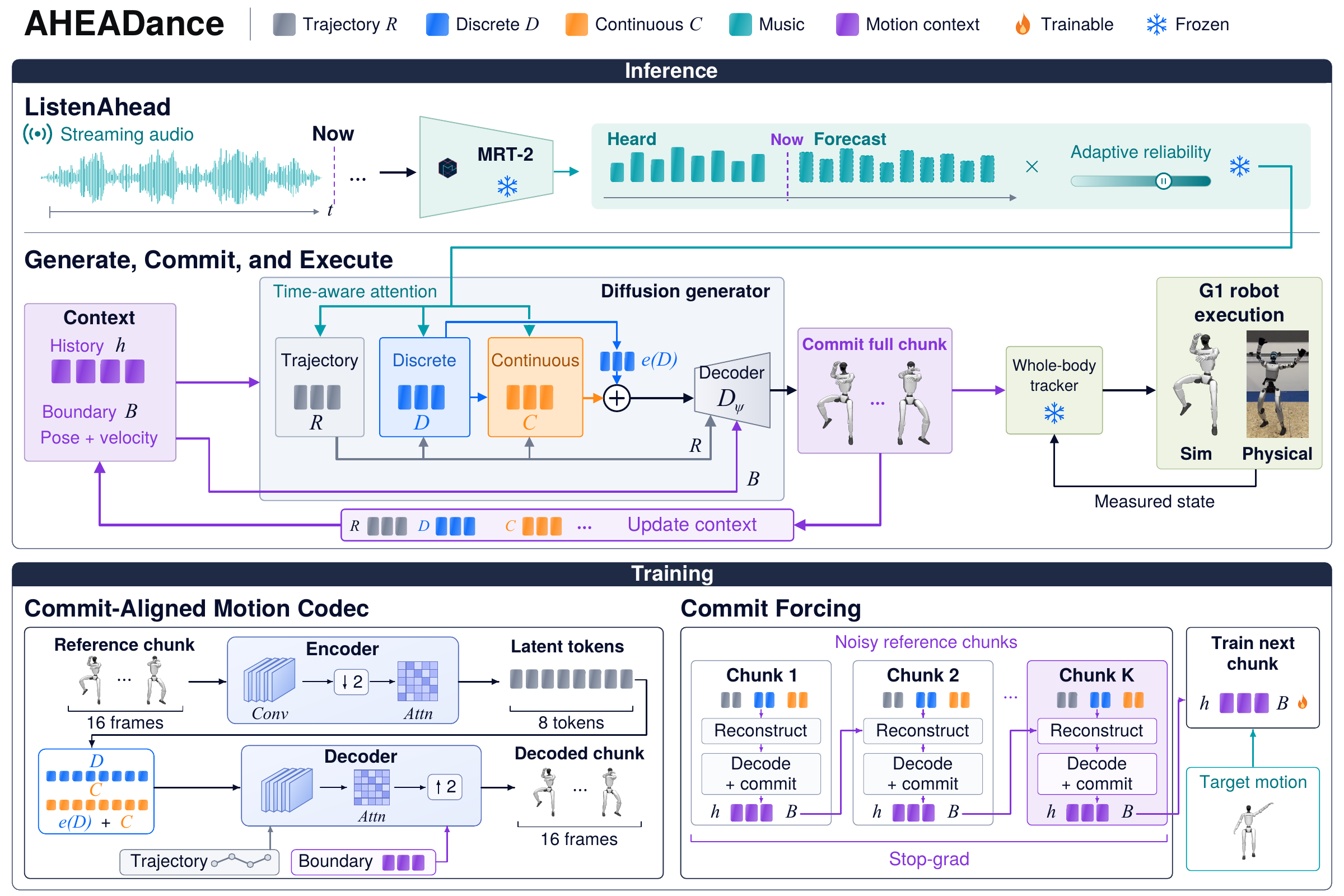}
\caption{Overview of \method{}. ListenAhead builds a time-aligned music
condition from heard audio and reliability-weighted forecast states. Each chunk is
generated in three stages (root trajectory, discrete codes, continuous corrections),
decoded once by the commit-aligned motion codec from the committed boundary,
committed, and tracked by a whole-body tracker.
Commit Forcing trains continuation from the model's own committed chunks.
Robot renderings are schematic, except in the Physical panel.}
\label{fig:overview}
\end{figure*}

We present \method{} (Figs.~\ref{fig:streaming_teaser} and~\ref{fig:overview}), which generates humanoid
reference motion chunk by chunk from streaming audio and executes it with
a whole-body tracking controller. Our contributions are:
\begin{itemize}
\setlength{\itemsep}{2pt}
\setlength{\parsep}{0pt}
\setlength{\topsep}{3pt}
\item \textbf{ListenAhead.} Each chunk is conditioned on heard music and
on short-horizon forecasts of a frozen music model~\cite{magenta2026mrt2},
aligned to their target times and weighted by their reliability. The
forecasts carry the timing of upcoming beats almost as well as the actual future,
and most of what the actual future carries about note onsets; a generator trained with them is
better synchronized
with the music and concentrates more motion energy at the song's beat
frequency.
\item \textbf{Commit Forcing.} Continuation is trained from the model's
own committed outputs: noise-corrupted reference chunks are reconstructed,
decoded and committed in order, and each commit updates both the latent
history and the decoded reference boundary that conditions the next chunk.
Low activity falls from 83.5\% under reference-history training to 10.3\%,
the level of real dance.
\item \textbf{A validated evaluation of music correspondence.} Three evaluators
test humanoid dance against its music: synchronization with ChoreoSync, a
detector of our own; song with MMR; and style with a style classifier. Each is
validated on real G1 dance it never saw. On songs unseen in training,
\method{} reaches the highest ChoreoSync offset accuracy (28.6\%) and MMR
(0.187, against at most 0.046) of all compared generators.
\end{itemize}
Code, evaluation scripts and the metric-validation protocol will be released
upon acceptance.

\section{Related Work}
\label{sec:related_work}

\subsection{Music-Driven and Streaming Dance}
AIST++/FACT~\cite{li2021aist}, Bailando~\cite{siyao2022bailando} and
Bailando++~\cite{siyao2023bailandopp} learn choreography through sequence
modeling and discrete motion representations, and
Danceba~\cite{fan2025danceba} adds gated rhythm-aware features for beat
alignment. EDGE~\cite{tseng2023edge} and
LODGE~\cite{li2024lodge} extend diffusion-based synthesis to long
soundtracks. ChoreoMamba uses bidirectional music conditioning for
long-sequence generation~\cite{tan2026choreomamba}.
FineDance~\cite{li2023finedance},
MEGADance~\cite{yang2025megadance}, and
SoulNet~\cite{li2025soulnet} further address fine-grained motion,
genre-aware generation, and music--motion correspondence.
Our focus is anticipatory continuation in robot coordinates as music arrives.

DiscoForcing~\cite{ji2026discoforcing} combines causal audio features
with diffusion forcing for streaming character motion, followed by
retargeting and tracking for humanoid execution, and shows responses to
music stopping qualitatively.
RoboPerform~\cite{li2026roboperform} instead maps audio-derived style
to humanoid actions through a teacher--student policy.
InfiniDanceRobot~\cite{zhang2026infinidancerobot} streams music features
into a latent diffusion-forcing planner that emits humanoid references, trains
a dedicated tracker, and closes the loop by refining the planner with
tracker rollouts and writing the executed state back into its history.
Table~\ref{tab:capabilities} distinguishes these interfaces from offline
choreography; unlike them, \method{} conditions on an explicit forecast of
future music representations from the frozen MRT-2 model~\cite{magenta2026mrt2}.

\begin{table}[t]
\papertablestyle
\caption{Interfaces of Music-Driven Methods}
\label{tab:capabilities}
\setlength{\tabcolsep}{2.4pt}
\begin{tabular}{@{}lccccc@{}}
\toprule
Method & Audio & Causal & Robot ref. & Own hist. & Forecast \\
\midrule
EDGE~\cite{tseng2023edge} & \checkmark & --- & --- & --- & --- \\
LODGE~\cite{li2024lodge} & \checkmark & --- & --- & --- & --- \\
DiscoForcing~\cite{ji2026discoforcing} & \checkmark & \checkmark & --- & --- & --- \\
InfiniDance~\cite{zhang2026infinidancerobot} & \checkmark & \checkmark & \checkmark & ---$^\dagger$ & --- \\
RoboPerform~\cite{li2026roboperform} & $\ast$ & \checkmark & --- & --- & --- \\
\method{} & \checkmark & \checkmark & \checkmark & \checkmark & \checkmark \\
\bottomrule
\end{tabular}
\tablenote{\checkmark: provided; ---: not described. Audio: raw audio input; $\ast$: described in the paper; the public release takes precomputed latents. Robot ref.: the generator itself outputs the robot reference motion that a tracking controller executes (DiscoForcing generates human motion and retargets it at run time). Own hist.: continuation is trained on the model's own outputs rather than on reference history. InfiniDance: InfiniDanceRobot. $^\dagger$Writes the executed robot state back into the history at inference.}
\end{table}

\subsection{Motion Representation and Generated Histories}
Discrete models such as T2M-GPT~\cite{zhang2023t2mgpt} and
MoMask~\cite{guo2024momask} learn motion vocabularies, while
MLD~\cite{chen2023mld} uses continuous latent diffusion.
DisCoRD~\cite{cho2025discord} connects discrete tokens to continuous
decoding, and DC-Motion~\cite{wang2026dcmotion} combines discrete
structure with continuous residuals. Our formulation builds on this
hybrid principle with sampled-code residual targets and explicit
trajectory and boundary conditions for robot-native chunks.

Learning from generated context also has substantial precedent.
Scheduled Sampling~\cite{bengio2015scheduled} and
Diffusion Forcing~\cite{chen2024diffusionforcing} address mismatched or
uncertain sequence context; DiscoForcing~\cite{ji2026discoforcing}
trains on noise-corrupted reference history and down-weights distant
history at inference. Self Forcing~\cite{huang2025selfforcing}
trains autoregressive video diffusion through self-rollouts and
sequence-level distribution matching.
MotionStreamer~\cite{xiao2025motionstreamer} combines causal motion
compression, diffusion autoregression, and Two-Forward training.
\cof{} constructs the history and the boundary state jointly: reconstructed
trajectory, codes and corrections enter committed history together, and their
decoded motion sets the next reference boundary.

\subsection{Humanoid Motion Execution}
Motion imitation and learned control provide reusable execution
layers. PHC~\cite{luo2023phc} studies reference tracking in simulated
avatars; ExBody2~\cite{ji2024exbody2},
BeyondMimic~\cite{liao2025beyondmimic}, and
SONIC~\cite{luo2026sonic} advance expressive humanoid control.
HARMON~\cite{jiang2025harmon} connects language to whole-body
humanoid generation. PhysDiff~\cite{yuan2023physdiff} incorporates
physical correction into denoising, while
PARC~\cite{xu2025parc} uses tracked motion to augment generator
training.
We retarget once and generate robot-native references for a fixed SONIC
controller, evaluating reference quality and tracking separately.

\subsection{Evaluating Generated Dance}
Generated dance is commonly scored by Fr\'echet distances of kinetic and
geometric motion features to real dance (FID$_k$, FID$_g$), by their
diversity and by the beat alignment score
(BAS)~\cite{siyao2022bailando,tseng2023edge}. Humanoid dance systems add
foot sliding~\cite{ji2026discoforcing}, foot contact and tracking
success~\cite{zhang2026infinidancerobot}, or retrieval and tracking
accuracy~\cite{li2026roboperform}. Learned music--motion matching scores
whether motion fits its song~\cite{li2025soulnet}, and movement descriptors
characterize style~\cite{turab2025dance}. We validate each music evaluator on
real G1 dance it never saw, and add a detector that estimates when motion
occurs relative to the music (Sec.~\ref{sec:evaluation_protocol}).

\section{Method}
\label{sec:method}

\subsection{Streaming Generation with Committed Motion}
\label{sec:streaming}
At decision time $\tau_t$, the generator has access to heard audio
$a_{\leq\tau_t}$, latent history $H_t$, and boundary state $B_t$
reconstructed from the last committed reference poses.
Let $M_t$ denote its music condition. Each decision generates one chunk
as a root trajectory $R_t$, discrete codes $D_t$, and continuous
corrections $C_t$ (Sec.~\ref{sec:dc}).

We use a commit-aligned motion codec: encoding, generation, and
commitment share the same fixed-length chunk. The codec compresses
reference body motion into latent vectors and decodes each generated
chunk once from its root trajectory and the preceding committed
boundary. Each complete chunk is committed before the next decision,
so committed motion is never decoded again.

The process has no prescribed terminal chunk: each commitment updates
a bounded motion history and boundary state before the next decision.
It therefore supports long-horizon generation without increasing the
motion-context length with elapsed time. The training data include standing and return-to-standing segments, so the generator, without any rule, stands when the heard music falls silent and dances again when it returns.
The evaluated horizons and implementation settings are given in
Appendix~\mbox{\ref{sec:protocol_appendix}}.

\subsection{Commit-Aligned Motion Codec}
\label{sec:dc}
Root trajectory represents position, height, and planar heading;
the body retains full orientation and native joint angles.
A new sampled code or incoming pose should change how a chunk is
reconstructed without changing its body-motion target. Our formulation
separates that target from the context used to decode it.
A chunk encoder maps body motion to latent vectors $z$ independently
of the generated history or decoder boundary. A learned codebook
contains structure embeddings $e_k$. For each vector,
\begin{equation}
D=\arg\min_k\|z-e_k\|^2,\qquad C=z-e_D .
\label{eq:latent}
\end{equation}
Codes supply motion prototypes and residuals supply continuous detail,
following the hybrid representation in~\cite{wang2026dcmotion}; we call
this combined body representation D+C.
Our decoder also receives the explicit root trajectory and current boundary:
\begin{equation}
x_t=\mathcal{D}\!\left(e(\widehat D_t)+\widehat C_t,\widehat R_t,B_t\right).
\label{eq:decode}
\end{equation}
Boundary information includes pose, joint velocity, root velocity, and
orientation change reconstructed from the last committed chunk, the
reference the tracker executes.
Body-only encoding keeps the target fixed as the preceding motion
changes; the decoder adapts it to the resulting pose and velocity.
The trajectory handles global placement separately from body structure.

\paragraph{Learning the codec}
The shared decoder learns from structure-plus-detail, structure-only,
and perturbed-residual reconstructions using robot-native pose,
forward-kinematic geometry, derivative, and boundary objectives.
Two consecutive chunks are decoded in sequence, as in streaming
commitment; the first reconstruction supplies the next boundary. Standard quantization and commitment losses
train the codebook, without predefined semantic or anatomical code labels.

\paragraph{Root, structure, then detail}
The generator samples the three parts of a chunk in sequence:
\begin{align}
p(R_t,D_t,C_t\mid H_t,B_t,M_t)
={}&p(R_t\mid H_t,B_t,M_t) \nonumber\\
&\cdot p(D_t\mid R_t,H_t,B_t,M_t) \nonumber\\
&\cdot p(C_t\mid D_t,R_t,H_t,B_t,M_t).
\label{eq:conditional}
\end{align}
Continuous diffusion first samples the trajectory, which fixes global
placement; categorical diffusion then samples the codes given the
trajectory, and conditional continuous diffusion samples the corrections
given both. All three stages share the history, boundary, and music
condition. Target positions within a chunk are modeled jointly, while
committed history remains fixed.

\paragraph{Detail conditioned on sampled structure}
The reference residual is tied to the reference code. Pairing it with a
different sampled code would shift the combined latent away from the
intended body target. We instead rebase the continuous target around
the code actually selected:
\begin{equation}
C_{\mathrm{target}}=z^*-e(\widehat D),
\qquad e(\widehat D)+C_{\mathrm{target}}=z^* .
\label{eq:rebase}
\end{equation}
Thus different code samples retain the same combined latent target
$z^*$ while inducing different correction targets. This is distinct
from learning a residual around a fixed reference code. Together with
boundary-independent encoding, it separates the intended body motion
from variations in sampled structure and preceding motion.
Training normalizes residuals with frozen training-set statistics.

\paragraph{Information access at the commitment scale}
Because the codec chunk is the commitment unit, the decoder attends jointly
over the chunk's whole generated plan, coordinating its entrance and exit,
before the chunk is committed. This reads no unheard waveform: causality is
imposed at the music decision boundary, not on plan variables already
generated inside the chunk. Across chunks, the committed pose and velocity and
cross-boundary derivative supervision provide continuity.

\subsection{ListenAhead Music Conditioning}
\label{sec:anticipation}
The chunk generated at decision time $\tau_t$ accompanies music that has not
yet been heard. ListenAhead supplies that music as a forecast. A frozen MRT-2
model predicts future hidden states from the observed audio
prefix~\cite{magenta2026mrt2}: 14 states at 25\,Hz whose target times reach about 0.5\,s past the audio cutoff
$\tau_t$, where the chunk begins, and so span nearly all of its 0.533\,s; each is
the mean of eight sampled continuations in hidden space. The forecast is computed once per decision and held fixed
while the chunk is sampled; Sec.~\ref{sec:music_results} measures what it
carries. ListenAhead turns it into a motion condition with two explicit
attributes: when a state is expected and how much the generator should rely
on it.

The music condition $M_t$ has three parts: the heard music, the forecasts, and
a MusicCoCa style embedding~\cite{caillon2025livemusic} of the trailing 10\,s.
Heard music enters through gated cross-attention over its states, style
through a scale and shift of the hidden states, and the forecasts through the
time-aware attention below.

\paragraph{When a forecast is expected}
Predicted states retain their physical time rather than becoming an unordered
future summary. For target motion position $i$ and forecast state $j$, let
$\Delta_{ij}$ be the difference between their times. Attention combines a
content score with a timing term $\beta(\Delta_{ij})$, a per-head linear map of
sinusoidal features of $\Delta_{ij}$ at 32 log-spaced frequencies:
\begin{align}
w_{ij}
&=\operatorname{softmax}_{j}
\left(q_i^\top k_j/\sqrt{d}+\beta(\Delta_{ij})\right), \\
u_i&=\sum_j w_{ij}\rho_jv_j ,
\label{eq:future_attention}
\end{align}
where invalid states are masked and $\rho_j$ is prediction reliability.
The timing term matches upcoming musical context to a motion position.

\paragraph{How much to rely on a forecast}
Forecasts are not equally good. The reliability $\rho_j\in[0,1]$ estimates how
much better forecast $j$ is than holding the last heard state. It is a ridge
regression on causal features: the lead time, how far the forecast departs
from the last heard state, the recent activity of the music, and the realized
skill of the two preceding forecasts. It is fitted once on the training split
and uses the audio prefix only. Reliability scales the forecast's contribution
after attention normalization. This placement lets the entire forecast pathway
weaken when predictions are uncertain, rather than redistributing a unit
attention mass among unreliable states.

\paragraph{Entering training at zero}
The forecast pathway is zero-initialized: its output projection and timing
term start at zero, so the forecasts contribute nothing at the start of
training and enter only as far as they reduce the loss. Masking all forecast
states and setting $\rho_j=0$ removes the pathway; this is the w/o-future model
of Sec.~\ref{sec:music_results}. The same music interface conditions
trajectory, discrete-code, and continuous-correction generation.

\subsection{Commit Forcing}
\label{sec:cof}
During generation, latent history and the incoming boundary both arise
from earlier model output. Replacing recorded latents with generated
ones while retaining a recorded boundary would describe two different
motion histories. \cof{} constructs both conditions from the same
decoded commitments.

\begin{figure*}[t]
\centering
\includegraphics[width=\textwidth]{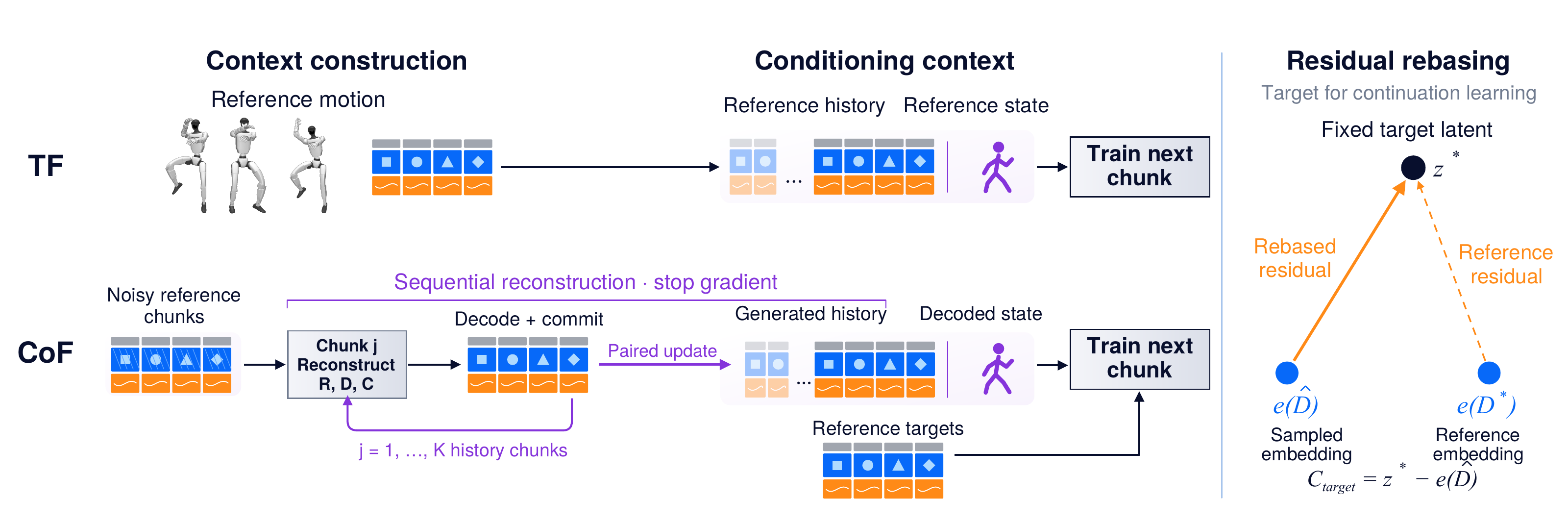}
\caption{\textbf{Commit-consistent continuation training.}
Reference-history training supplies recorded contexts. Commit Forcing
reconstructs a sequence of $K$ perturbed reference chunks, decodes and
commits each complete chunk, and carries the updated context into the
next reconstruction. Each update pairs generated history with the
boundary state of that same decoded motion. This history construction
stops gradients; the current continuation is supervised with reference
targets. The inset uses the shared code-relative target rule
of Eq.~\ref{eq:rebase}.}
\label{fig:cof}
\end{figure*}

After reference-history warmup, the current generator reconstructs
trajectory, codes, and continuous corrections from noise-corrupted
reference chunks with one denoising step
(Appendix~\mbox{\ref{sec:implementation}}). Starting from the context at the beginning of the
history window, it processes chunks $j=1,\ldots,K$ in sequence.
Each reconstruction is decoded and committed before the next one:
\begin{equation}
(H_j,B_j)
=\mathcal{U}(H_{j-1},B_{j-1},\widehat R_j,\widehat D_j,\widehat C_j).
\label{eq:joint_context}
\end{equation}
Here $\mathcal{U}$ forms $H_j$ by appending generated latent components
and reconstructs $B_j$ from their decoded terminal poses.
The next history reconstruction receives this updated pair, so changes
in generated motion reach both the latent context and the starting state.
After $K$ updates, $(H_K,B_K)$ conditions continuation learning.

History reconstruction stops gradients; code prediction, continuous
denoising, and decoded-motion objectives supervise the next chunk.
The residual target rule in Sec.~\ref{sec:dc} is shared by history
reconstruction and continuation supervision.

\subsection{Robot Execution}
\label{sec:robot_execution}
SONIC~\cite{luo2026sonic} tracks the generated motion in simulation.
The interface converts joint, orientation, and timing conventions.
The generator updates its history and boundary state from committed motion;
SONIC uses measured state for feedback control.
The music scores in Sec.~\ref{sec:experiments} are computed on the generated
reference motion, before execution.

\section{Experiments}
\label{sec:experiments}

Table~\ref{tab:baselines} compares generators, Table~\ref{tab:ablation}
isolates ListenAhead and \cof{}, and Table~\ref{tab:admission} validates the
evaluators on real G1 dance.

\subsection{Setup and Metrics}
\label{sec:evaluation_protocol}

\begin{table}[t]
\papertablestyle
\caption{Comparison on 42 Unseen Songs}
\label{tab:baselines}
\setlength{\tabcolsep}{3pt}
\begin{tabular}{@{}l ccccc@{}}
\toprule
Method & FID$_k\downarrow$ & Div$_k$ & Sync$\uparrow$ & MMR$\uparrow$ & Tempo$\uparrow$ \\
\midrule
Real dance & -- & 6.22 & 69.0 & 0.245 & 80.2 \\
\midrule
\multicolumn{6}{@{}l}{\emph{Online (no future music)}} \\
\rowcolor{gray!15}\method{} & \textbf{18.8} & 2.22 & \textbf{28.6} & \textbf{0.187} & 57.4 \\
Bailando++, strict & 65.2 & 0.59 & 11.9 & $-$0.010 & 48.1 \\
Danceba, strict & 44.8 & 4.52 & 9.5 & $-$0.003 & 46.6 \\
MEGADance, strict & 23.7 & 3.77 & 11.9 & $-$0.011 & 49.4 \\
\midrule
\multicolumn{6}{@{}l}{\emph{Offline}} \\
EDGE & 27.6 & 2.22 & 27.8 & 0.046 & \textbf{58.8} \\
Lodge & 44.1 & 1.70 & 14.3 & 0.044 & 46.2 \\
MEGADance & 47.1 & 2.46 & 10.3 & 0.008 & 46.8 \\
Bailando++ & 77.8 & 0.27 & 4.8 & $-$0.007 & 50.2 \\
Danceba & 61.0 & 1.79 & 11.9 & 0.007 & 53.4 \\
\bottomrule
\end{tabular}
\tablenote{42 AIOZ-GDANCE test songs whose audio occurs in no model's training data. All motion is generated G1 reference motion, before execution; every model of a song is scored on the same frames. Sync: ChoreoSync offset accuracy (\%, chance 11.1). Other scores as in Table~\ref{tab:ablation}; FID$_k$ against real dance of these songs. No Style column: our style classifiers cover only the FineDance and AIST++ style labels; Appendix~\ref{sec:style_unheard} compares style on FineDance songs. Bailando++ and Danceba: one sample per song; others three. Online, strict: the model runs on causal input. Baseline details in Appendix~\mbox{\ref{sec:baseline_details}}. Bold: best generator in the column.}
\end{table}

\begin{table*}[t]
\papertablestyle
\caption{Training Ablations on 721 Held-Out Sequences}
\label{tab:ablation}
\setlength{\tabcolsep}{4pt}
\begin{tabular}{@{}lcccc cc ccc@{}}
\toprule
& \multicolumn{4}{c}{Music} & \multicolumn{2}{c}{Continuation} & \multicolumn{3}{c}{Motion distribution} \\
\cmidrule(lr){2-5}\cmidrule(lr){6-7}\cmidrule(l){8-10}
Setting & ChoreoSync$\uparrow$ & MMR$\uparrow$ & BAS$\uparrow$ & Tempo$\uparrow$ & Low act.$\downarrow$ & Span$\uparrow$ & FID$_k\downarrow$ & FID$_g\downarrow$ & Div$_k$ \\
\midrule
Real dance & 45.2 & 0.184 & 0.247 & 71.3 & 9.7 & 0.922 & -- & -- & 6.28 \\
\midrule
\rowcolor{gray!15}\method{} & \textbf{24.2} & \textbf{0.126} & 0.262 & \textbf{56.8} & \textbf{10.3} & \textbf{0.732} & \textbf{16.7} & \textbf{46.3} & 2.94 \\
w/o future channel & 17.4 & 0.054 & 0.268 & 51.3 & 17.5 & 0.604 & 21.3 & 74.1 & 3.11 \\
w/o forecasts (last heard state) & 15.8 & 0.047 & 0.265 & 48.3 & 10.8 & 0.636 & 23.2 & 71.1 & 2.55 \\
Teacher Forcing & 18.4 & 0.023 & 0.238 & 47.6 & 83.5 & 0.492 & 64.6 & 140.6 & 1.53 \\
CoF w/o boundary closure & 18.5 & 0.029 & 0.294 & 53.3 & 72.5 & 0.509 & 45.1 & 123.8 & 1.77 \\
\bottomrule
\end{tabular}
\tablenote{Each row is a separately trained model (three sampling seeds); test and validation sequences of at least 8\,s, up to 32.8\,s (224 songs). w/o forecasts: the future channel holds the last heard music state. ChoreoSync: offset estimated within 0.2\,s of the true alignment (\%, chance 11.1). MMR: margin of a motion window's similarity to its own music over other songs' music at the same time. BAS: beat alignment score. Tempo: tempo salience (\%, chance 50). Low act.: 4\,s windows below the 10th percentile of training motion (\%). Span: joint range (rad). FID$_k$, FID$_g$ and Div$_k$ (descriptive): 5\,s windows, per dataset against the real dance of these sequences (Appendix~\mbox{\ref{sec:metrics_appendix}}). Bold: best generator in the column; BAS is not ranked.}
\end{table*}

\begin{table}[t]
\papertablestyle
\caption{Score Validation on Real G1 Dance}
\label{tab:admission}
\setlength{\tabcolsep}{1.2pt}
\begin{tabular}{@{}lcccccc@{}}
\toprule
Score & AIOZ & FineDance & AIST++ & CoMPAS3D & All & Chance \\
Sequences & 506 & 103 & 95 & 17 & 721 & \\
\midrule
ChoreoSync (\%) & 46.2 & 26.2 & 53.7 & 82.4 & 45.2 & 11.1 \\
MMR (\%) & 68.4 & 68.7 & 67.6 & 80.9 & 68.6 & 50 \\
Tempo (\%) & 71.2 & 68.1 & 79.3 & 45.9 & 71.3 & 50 \\
Style (\%) & -- & 84.2 & 65.3 & -- & -- & 33\,/\,10 \\
BAS, own music & 0.247 & 0.216 & 0.283 & 0.212 & 0.247 & -- \\
BAS, other songs & 0.244 & 0.214 & 0.250 & 0.250 & 0.241 & -- \\
\bottomrule
\end{tabular}
\tablenote{Test and validation sequences that the evaluators were not trained on. MMR: share of comparisons in which a window's own music scores above another song's music at the same time. Style: 76 labelled FineDance sequences and the 95 AIST++ sequences, classifiers that never saw the song; chance 33\% on FineDance (three classes), 10\% on AIST++ (ten genres). Tempo salience on CoMPAS3D stays at chance.}
\end{table}

We use dance motion from six public datasets retargeted to our test platform,
the Unitree G1 humanoid:
AIOZ-GDANCE~\cite{le2023aiozgdance}, FineDance~\cite{li2023finedance},
CoMPAS3D~\cite{burkanova2025compas3d}, AIST++~\cite{li2021aist},
ChoreoMaster~\cite{chen2021choreomaster} and
OpenDance~\cite{zhang2026opendance}, as released in OMG-Data~\cite{huang2026omg}
with that release's official splits (61.7\,h of training motion).

\begin{figure*}[t]
\centering
\includegraphics[width=\textwidth]{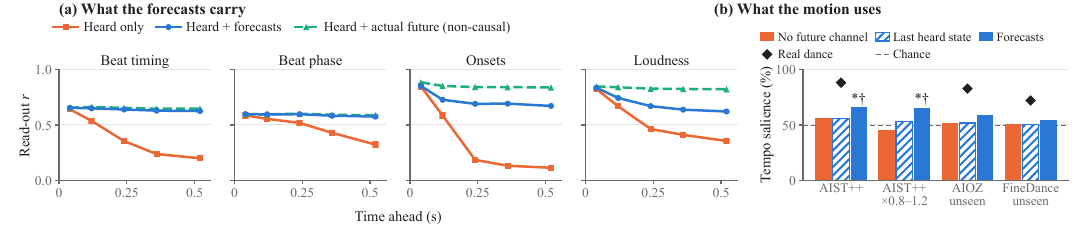}
\caption{ListenAhead. (a) What the forecasts carry: cross-validated linear
read-outs (Pearson $r$ on held-out songs; five song folds, 235 songs) of music
descriptors, by how far ahead, from MRT-2 states of the heard music, the heard
music with the forecasts, or the heard music with the actual future (a
non-causal upper bound). (b) What the motion uses: tempo salience of three
generators (without the future channel, with it holding the last heard music
state, and with the forecasts) on AIST++ music (60 pieces), ten of these
pieces time-stretched to five speeds between 0.8$\times$ and 1.2$\times$, and
songs unseen in training (AIOZ-GDANCE 42, FineDance 30). $^*$, $^\dagger$: the
95\% CI of the paired difference over songs from the generator without the
future channel ($^*$) or with the last heard state ($^\dagger$) excludes zero
(t interval for the ten time-stretched pieces). Diamonds: real dance of the
same songs (51 AIST++ pieces and 27 FineDance songs; none for time-stretched
music); dashed line: chance.}
\label{fig:listenahead}
\end{figure*}

\textbf{Evaluation sets.} Ablations use 721 sequences (224 songs) of at least
8\,s from the test and validation splits, none of which ChoreoSync was trained
on: AIOZ-GDANCE 506, FineDance 103, AIST++ 95 and CoMPAS3D 17. Generators are
compared on 42 of these songs: by audio fingerprint, their audio occurs
neither in any generator's training data nor in that of ChoreoSync or MMR.

\textbf{Music scores.} Three evaluators test how generated dance corresponds
to its music: in synchronization, in song and in style.

\emph{Synchronization.} ChoreoSync, our contrastive motion--music detector
trained on G1 dance, estimates the music offset at which a clip's motion and
music agree best; offset accuracy, the share of clips whose estimate lies
within 0.2\,s of the true alignment, measures whether movement is aligned in
time with its own music (Appendix~\mbox{\ref{sec:score_validation}}).

\emph{Song.} MMR, the music--motion retrieval model of
SoulNet~\cite{li2025soulnet} retrained on G1 motion, scores the margin of a
motion window's similarity to its own music over other songs' music at the
same time, testing whether the window matches its song; every sequence is
scored only by evaluators not trained on it
(Appendix~\mbox{\ref{sec:score_validation}}).

\emph{Style.} A classifier over movement descriptors~\cite{turab2025dance}
labels FineDance dances Classic, Folk or Street, and a second labels AIST++
dances by their ten genres; style accuracy is the share of samples classified
as their song's style, each song scored by a classifier fitted without it.
Style following is the rate at which a dance is classified as its own song's
style minus the mean rate at which it is classified as the style of each other
song, so a dance whose style ignores its music scores zero.

The evaluators are not used to train the generator. We also report the
common beat alignment score (BAS)~\cite{siyao2022bailando}.

\textbf{Tempo salience.} A rule-based score of beat-frequency energy: the share
of comparisons in which a dance concentrates more motion energy at its own
song's beat frequency than at that of a rival song whose tempo differs by at
least 8\%.

\textbf{Validation.} Each evaluator is validated on real G1 dance
(Table~\ref{tab:admission}; Appendix~\mbox{\ref{sec:score_validation}}).
ChoreoSync estimates the offset within 0.2\,s for 45.2\% [39.5, 50.9] of the
721 sequences (chance 11.1\%), and MMR prefers a dance's own music in 68.6\%
[65.9, 71.4] of comparisons (chance 50\%). BAS does not distinguish real dance
on its own music from real dance on other songs' music (0.247 vs.\ 0.241).

\textbf{Motion.} Joint span and low activity characterize sustained movement.
Motion distributions are compared with the kinetic and geometric Fr\'echet
distances (FID$_k$, FID$_g$) and kinetic diversity
(Div$_k$)~\cite{siyao2022bailando}; windows and normalization are given in
Appendix~\mbox{\ref{sec:metrics_appendix}}. Appendix~\mbox{\ref{sec:laban_appendix}} adds
FID$_{\mathrm{LMA}}$ and Laban movement qualities.
Paired differences use the compact notation
\emph{estimate [95\% confidence interval]}, with the complete model minus the
compared model.

\subsection{Comparison with Other Generators}
\label{sec:baselines}
\label{sec:roboperform}

We compare \method{} with EDGE~\cite{tseng2023edge},
Lodge~\cite{li2024lodge}, MEGADance~\cite{yang2025megadance},
Bailando++~\cite{siyao2023bailandopp} and Danceba~\cite{fan2025danceba}. To
compare them under the same causal access, we also run Bailando++, Danceba and
MEGADance strictly online, with music features recomputed at each step from
the heard audio only and token decoders that see no future tokens
(Appendix~\mbox{\ref{sec:baseline_details}}).

On 42 songs unseen in training, \method{} has the lowest FID$_k$ and the
highest MMR of all generators (Table~\ref{tab:baselines}): FID$_k$ 18.8
against at least 23.7, and MMR 0.187 against at most 0.046, with the 95\% CI
of every difference excluding zero. Its ChoreoSync offset accuracy, 28.6\%, is
also the highest. RoboPerform's release takes music latents precomputed
for its training clips and cannot be given other music, and DiscoForcing's
weights are not released (our reproduction is reported in
Appendix~\mbox{\ref{sec:baseline_details}}), so we compare their interfaces
(Table~\ref{tab:capabilities}). On 23 FineDance songs that \method{} never
heard, each scored by a classifier fitted without it, its style following is
\resultci{32.6}{22.5}{40.9} points (real dance 47.4): higher than that of every
generator given no genre and every strictly online variant, not significantly
different from that of Lodge, and lower than that of offline MEGADance, both of
which receive each song's genre and were trained on FineDance
(Appendix~\mbox{\ref{sec:style_unheard}}).

\subsection{Music Alignment with ListenAhead}
\label{sec:music_results}

Trained without the future channel, the generator's ChoreoSync offset accuracy
is \resultci{6.8}{2.4}{11.2} points lower (17.4\% vs.\ 24.2\%) and its tempo
salience \resultci{5.5}{2.3}{8.7} points lower (51.3\% vs.\ 56.8\%;
Table~\ref{tab:ablation}). The future channel also lets tempo follow the music when it is played 0.8--1.2$\times$ faster or
slower: tempo salience rises from 45.0\% to 65.7\%
(\resultci{20.7}{14.3}{27.1}; higher for all ten pieces;
Fig.~\ref{fig:listenahead}b). With the forecast slots filled by the last heard
music state instead, the gain disappears at the original tempo: tempo salience on the 60 AIST++ pieces is
\resultci{10.6}{5.8}{15.5} points lower than with the forecasts and no higher
than without the future channel, and ChoreoSync offset accuracy is
\resultci{8.4}{3.3}{13.5} points lower (15.8\% vs.\ 24.2\%;
Table~\ref{tab:ablation}). At 0.8--1.2$\times$ it is
\resultci{12.3}{6.4}{18.2} points lower, keeping 40\% of the gain, so the gain
comes mainly from what the forecasts predict rather than from an extra
time-aligned input.

\textbf{What the forecasts encode.} We read music descriptors from MRT-2
states with cross-validated linear read-outs (PCA and ridge regression) on
235 songs from FineDance, AIST++, AIOZ-GDANCE, PhantomDance and 16 other
tracks (Fig.~\ref{fig:listenahead}a). The forecasts anticipate rhythm: on average
over 0.04--0.52\,s ahead, they provide 95\% of the gain that the actual future
gives for the timing of upcoming beats, 96\% for beat phase and 74\% for note
onsets, and help least for changes in loudness. A causal beat tracker that
extrapolates the heard beats recovers 46\% of the forecasts' gain for beat
timing, 58\% for beat phase and 12\% for onsets. At 0.52\,s ahead, beat timing
is read with $r=0.63$ from the forecasts (actual future 0.65) and $r=0.20$
from the heard music alone.

\subsection{Sustained Motion with Commit Forcing}
\label{sec:continuation_results}
\cof{} sustains movement relative to reference-history training
(Table~\ref{tab:ablation}). Low activity falls from 83.5\% to 10.3\%, the
level of real dance (9.7\%; \resultci{-73.1}{-76.2}{-69.9} points). Later in
continuation the gap widens: from 8.8 to 32.8\,s, on the 284 sequences that
reach 32.8\,s, low activity falls from 98.5\% to 8.4\%. \cof{} also moves motion
closer to real dance: FID$_k$ falls from 64.6 to 16.7
(\resultci{-47.9}{-53.6}{-42.6}) and FID$_g$ from 140.6 to 46.3
(\resultci{-94.3}{-104.7}{-87.6}).
With the decoded reference boundary held at the reference (latent history
still self-generated), low activity rises to 72.5\%
(\resultci{-62.1}{-65.2}{-59.0} points for the complete model) and FID$_k$ to
45.1 (\resultci{-28.4}{-33.7}{-24.3}), so once the history is self-generated,
the decoded boundary must be generated too.

\subsection{Execution on the Robot}
\label{sec:execution_results}
\begin{figure}[t]
\centering
\includegraphics[width=\columnwidth]{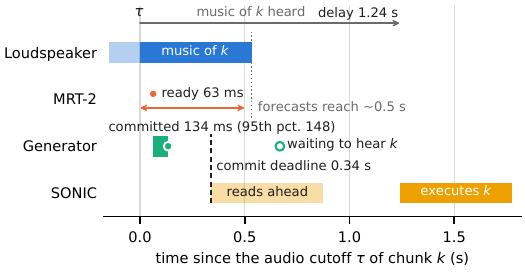}
\caption{One chunk on the robot, relative to its audio cutoff $\tau$: medians
over the 12 physical runs. Dashed: commit
deadline (1.24\,s delay minus the 0.9\,s preview); open circle: commit time if
each chunk's music were heard before generating it.}
\label{fig:robot_timeline}
\end{figure}
\begin{figure*}[t]
\centering
\includegraphics[width=\textwidth]{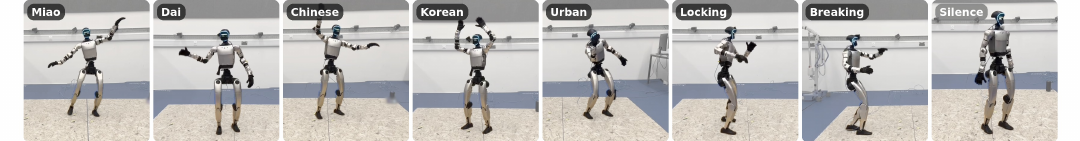}
\caption{The physical Unitree G1 dancing to songs of seven styles and standing after the music stops (backgrounds lightened).}
\label{fig:g1_styles}
\end{figure*}
\method{} generates reference motion faster than playback: on the robot, a
16-frame chunk (0.533\,s at 30\,Hz) is committed 0.13\,s after its audio cutoff
(0.15\,s at the 95th percentile), forecasting included
(Fig.~\ref{fig:robot_timeline}; Appendix~\mbox{\ref{sec:timing_notes}}). Because the SONIC tracker reads
the reference about 0.9\,s ahead, we execute on the robot with a 1.24\,s delay; waiting to
hear each chunk's music before generating it would add the chunk length, 0.53\,s.
With the official low-latency SONIC (0.32\,s preview), the same generator runs
all 8 programmes in real time in simulation with a 0.8\,s delay, which waiting
to hear each chunk's music could not meet
(Appendix~\mbox{\ref{sec:interruption_details}}). SONIC executes the generated
references in simulation successfully (no reset, pelvis height within
0.25\,m) in 99.5\% of 369 trials on the test split.

Dancing to music includes coming to rest when it stops
(Sec.~\ref{sec:method}). We interrupt songs with 60\,s of silence in 28 programmes, each paired with a control that continues the music from the same generator state
(Appendix~\mbox{\ref{sec:interruption_details}}). \method{} stops and dances
again in every programme; executed by SONIC in simulation, it is near standing
a median 2.32\,s after the music stops on the 16 songs from the test split.
Of the other generators, all trained on dance alone, Bailando++ stops and dances
again in 3 programmes, Danceba in 2, and EDGE, Lodge, MEGADance and the strictly
online variants in none. On a physical Unitree G1, the music is streamed from audio files
and played through a loudspeaker as it arrives, so the robot moves about
1.2\,s behind the music the audience hears. Near-standing followed each
of the 20 music stops and never occurred while the robot danced, and sustained
dancing followed each of the 20 starts, a median 1.75\,s and 0.85\,s later in the
stream (about 3.0\,s and 2.1\,s for the audience); no run fell, and joints track
the reference with a median error of 5.5$^\circ$ (Fig.~\ref{fig:g1_styles}; supplementary video;
Appendix~\mbox{\ref{sec:physical_runs}}).

\section{Discussion}
\label{sec:evidence_scope}

The forecasts carry the timing of the next beats almost as well as the actual
future, and most of what the actual future carries about note onsets; the
generator converts them into better synchronization and more motion energy at
the music's beat frequency. Turning anticipated beats into the timing of
movement accents is the most direct next step (Sec.~\ref{sec:music_results}),
followed by a perceptual study of how audiences judge the dance.

\method{}'s kinetic diversity still remains below that of real dance. Forecasts
anticipate regular structure such as beats, but not an abrupt change such as a
sudden stop, to which the robot responds afterwards (about 1\,s in simulation).
The music scores are measured on the generated references, which a
general-purpose tracker executes after a fixed delivery delay; a tracker trained
for dance could preserve more of their beat-level detail. The generator does
not receive the measured robot state, and it runs off-board on a workstation;
feeding the state back and running on the robot are left to future work.

\section{Conclusion}
\method{} generates robot-native dance from streaming audio under causal
music access and irrevocable commitments. A future-aligned music channel
anticipates beat timing and improves synchronization and beat-frequency motion
over a generator trained without it, a gain that the last heard music state
alone does not provide, and training continuation from the model's
own committed chunks keeps low activity at the level of real dance. ChoreoSync
and MMR, validated on real G1 dance, show that \method{} follows unseen songs
more closely than other generators, and on FineDance songs it never heard, its
style follows the music more closely than that of any generator not given the
genre.

\FloatBarrier
\FloatBarrier
\bibliographystyle{IEEEtran}
\bibliography{Reference}
\appendices
\section{Supplementary Results and Notes}
\label{sec:protocol_appendix}
\renewcommand{\theparagraph}{\thesection-\alph{paragraph}}

\paragraph{Evaluation settings}
The six datasets follow the official splits of the OMG-Data
release~\cite{huang2026omg} (\url{https://huggingface.co/datasets/THU-MARS/OMG-Data},
revision 6e0dfbc): 61.7\,h of training, 6.9\,h of validation and
7.5\,h of test motion. Training motion comprises AIOZ-GDANCE 48.5\,h, FineDance
5.5\,h, CoMPAS3D 4.5\,h, AIST++ 2.4\,h, ChoreoMaster 0.9\,h and OpenDance
0.04\,h. For FineDance, this split differs from the dataset's own test list:
the audio of 15 of its 18 test songs occurs in the training data, which applies
equally to every row of Table~\ref{tab:ablation}.
The 721 sequences are the test and validation sequences of at least 8\,s,
none of which was used to train ChoreoSync or the MMR evaluator that scores it.
By audio fingerprint, the music of 42 of them occurs in ChoreoSync's training
data and that of 101 in MMR's, through other dances to the same songs; every
row of Table~\ref{tab:ablation} is scored on the same sequences. Without these
sequences, the ChoreoSync differences of Sec.~\ref{sec:music_results} are
\resultci{7.0}{2.4}{11.6} and \resultci{8.7}{3.3}{13.9} points, and \method{}'s
MMR stays above that of every ablation in Table~\ref{tab:ablation}, each 95\%
CI excluding zero. The 721 sequences run up to
32.8\,s,
with three sampling seeds (1234, 2345, 3456) and ten sampling steps. Low
activity from 8.8 to 32.8\,s uses the 284 that reach 32.8\,s (113 songs).
The 42 AIOZ-GDANCE songs are a subset of these sequences, screened with a
landmark audio fingerprint against \method{}'s training data; the screen removed
two of 44 songs. Every model of a song is cut to the same frames, inside the song's real audio.

\paragraph{Implementation}
\label{sec:implementation}
\label{sec:models_appendix}
\method{} is trained from scratch (seed 1234) with a static standing initial
boundary. The codec (10.7\,M parameters) encodes two frames per latent vector
(16 dimensions; codebook of 512 entries). The generator (619\,M parameters;
width 768, 12 layers with 12 heads, 9 diffusion layers) conditions on 64
history tokens (128 frames, 4.27\,s) and generates eight tokens (one 16-frame
chunk) per decision with ten sampling steps per stage. MRT-2 (small) samples its
eight continuations at temperature 1.3 and top-$k$ 40. The reliability
regression is clipped to $[0,1]$. \cof{} reconstructs and commits the
$K=8$ chunks (4.27\,s) preceding each training window. Training uses reference
histories first and then switches to reconstructed histories, without a gradual
mixing schedule. Each source window draws one shifted logit-normal noise strength
(shift 0.6), shared across its successive chunks; each chunk is reconstructed with
one denoising step, and the history is built without gradient propagation. The generator is trained on four GPUs with 1{,}536 target chunks
per update; generation runs off-board on a workstation with separate GPUs for
the music model and the generator.

\paragraph{Score validation}
\label{sec:score_validation}
A music--motion score supports claims about musical correspondence only if it
prefers a dance's own music to other music. Our MMR is checked this way on
sequences its scoring evaluator was not trained on and, unlike SoulNet's
evaluator~\cite{li2025soulnet}, is never used to train the generator.

ChoreoSync has two Transformer encoders, $f_m$ and $f_a$, that map a 4\,s
window of G1 motion features and a 4\,s window of frozen MuQ music
features~\cite{zhu2025muq}
to unit embeddings. A symmetric contrastive loss takes the time-aligned
music window as the positive and, as negatives, windows of the same song
shifted by at least 0.15\,s plus other songs in the batch, so that the
embedding must encode when movement happens relative to the music rather
than which song plays. A clip is scored by the offset curve
\begin{equation}
c(\delta)=\frac{1}{|\mathcal{W}|}\sum_{t\in\mathcal{W}}
f_m(x_t)^{\top} f_a(a_{t+\delta}),\qquad \delta\in[-2,2]\,\mathrm{s},
\label{eq:choreosync}
\end{equation}
over motion windows $x_t$ starting every 1\,s and music windows
$a_{t+\delta}$ shifted in 0.05\,s steps.
The clip's offset estimate is the peak $\hat\delta=\arg\max_\delta c(\delta)$. ChoreoSync is trained on the training splits of the same datasets as
the generator and on real dance retargeted with GMR~\cite{araujo2025gmr}, so
that one detector scores both retargeting routes.
On generated motion, ChoreoSync follows known shifts: shifting \method{}'s
dance against its music by $\pm$0.25, $\pm$0.5 or $\pm$1\,s (679 sequences of
207 songs it never heard) moves the estimated offset by the same amount (median
error at most 0.10\,s), and accuracy against the unshifted alignment falls from
24.0\% to 8.6--13.7\% (chance 11.1\%) for shifts of 0.5\,s or more.

MMR follows the music--motion retrieval module of SoulNet~\cite{li2025soulnet},
a contrastive encoder--decoder in the TEMOS/TMR
style~\cite{petrovich2022temos,petrovich2023tmr} over Jukebox
features~\cite{dhariwal2020jukebox}, trained here on G1 motion. It has three
evaluators, each trained on two thirds of the FineDance, AIST++
and PhantomDance songs and on all training clips of AIOZ-GDANCE and CoMPAS3D of
at least 5\,s. FineDance and AIST++ sequences are scored by the evaluator that
holds them out, AIOZ-GDANCE and CoMPAS3D sequences by the mean of the three. MMR's
own-music rates on real dance have 95\% CIs [65.4, 71.3], [63.5, 73.9],
[54.9, 79.4] and [74.4, 87.4]\% on AIOZ-GDANCE, FineDance, AIST++ and CoMPAS3D.
Every style score, of real and of generated dance, uses a classifier refitted
without the scored song's fold and without songs that share its audio. BAS on real dance, own music minus other songs' music, is
\resultci{0.006}{-0.002}{0.014} on the 721 sequences.

\paragraph{Chunk representation}
\label{sec:representation_appendix}
\begin{table*}[t]
\papertablestyle
\caption{Chunk Representation on the Same 721 Sequences}
\label{tab:dc}
\setlength{\tabcolsep}{4pt}
\begin{tabular}{@{}lcccc cc ccc@{}}
\toprule
Setting & ChoreoSync$\uparrow$ & MMR$\uparrow$ & BAS$\uparrow$ & Tempo$\uparrow$ & Low act.$\downarrow$ & Span$\uparrow$ & FID$_k\downarrow$ & FID$_g\downarrow$ & Div$_k$ \\
\midrule
\rowcolor{gray!15}\method{} (D+C) & 24.2 & \textbf{0.126} & 0.262 & 56.8 & \textbf{10.3} & \textbf{0.732} & \textbf{16.7} & \textbf{46.3} & 2.94 \\
Discrete only (D) & 20.4 & 0.078 & 0.283 & 51.4 & 15.1 & 0.676 & 26.2 & 64.9 & 2.44 \\
Continuous only (C) & \textbf{28.0} & 0.054 & 0.270 & \textbf{57.4} & 34.6 & 0.589 & 28.5 & 81.3 & 2.60 \\
\bottomrule
\end{tabular}
\tablenote{Scores and sequences as in Table~\ref{tab:ablation}. Bold: best in the column; BAS is not ranked.}
\end{table*}

D+C is more precise than discrete-only generation and moves more than
continuous-only generation (Table~\ref{tab:dc}). Codes alone cannot place a pose
precisely: on the 123 sequences of the test split, decoding encoded motion chunk
by chunk from its own decoded boundary, as in streaming, gives 7.2$^\circ$ mean
joint error from codes alone, against 2.9$^\circ$ with residuals (higher for 122
of 123 sequences). In simulation on the same test split, SONIC tracks
discrete-only generation with \resultci{1.24}{0.97}{1.50}$^\circ$ more joint
error and \resultci{4.3}{2.5}{5.6}\,mm more position error than D+C (369 trials
per model). Continuous-only generation is tracked as accurately as D+C, and its
low activity is 34.6\% against 10.3\%.

\paragraph{Metrics}
\label{sec:metrics_appendix}
MMR takes its margin over the mean of other songs' music, in windows at 5, 10,
\ldots, 25\,s; it is a score, not an accuracy percentage.
BAS uses nearest music-to-motion beat matching on G1 keypoint speeds
with a three-frame Gaussian tolerance at 30\,Hz: music beats come from a
beat tracker, motion beats from local minima of the smoothed mean keypoint
speed.
Tempo salience runs from 1\,s to the end of a clip.
Joint span is the mean 95th--5th percentile joint range;
low activity counts 4-second windows below the 10th percentile of
training-motion activity, so real dance is near 10\%.
FID$_k$, FID$_g$ and Div$_k$ use 5\,s windows starting at 5, 10, \ldots, 25\,s
(a sequence of 8--10\,s uses the window at 0\,s). Kinetic features average
velocity and acceleration over $\pm 2$ frames, as in the Bailando code, and
features are normalized by the real dance of the same sequences. Fr\'echet
distances are taken against those sequences and averaged over datasets in
Table~\ref{tab:ablation}; Div$_k$ is the mean pairwise distance of normalized
kinetic windows.

\paragraph{Laban movement qualities}
\label{sec:laban_appendix}
\begin{table*}[t]
\papertablestyle
\caption{FID$_{\mathrm{LMA}}$ and Laban Movement Qualities}
\label{tab:laban}
\setlength{\tabcolsep}{3.0pt}
\begin{tabular}{@{}lcc cccccc@{}}
\toprule
& \multicolumn{2}{c}{Real dance} & \multicolumn{6}{c}{721 sequences} \\
\cmidrule(lr){2-3}\cmidrule(l){4-9}
Quality & Split & 1\,s loop & \method{} & w/o future & TF & w/o closure & D & C \\
\midrule
FID$_{\mathrm{LMA}}$ & 15.2 & 1443.7 & 33.6 & 41.7$^+$ & 103.3$^+$ & 76.7$^+$ & 47.1$^+$ & 52.3$^+$ \\
\bottomrule
\end{tabular}

\vspace{6pt}
\begin{tabular}{@{}lccccccccc@{}}
\toprule
& & \multicolumn{3}{c}{Online, strict} & \multicolumn{5}{c}{Offline, released} \\
\cmidrule(lr){3-5}\cmidrule(l){6-10}
42 songs & \method{} & Bailando++ & Danceba & MEGADance & EDGE & Lodge & MEGADance & Bailando++ & Danceba \\
\midrule
FID$_{\mathrm{LMA}}$ & 35.0 & 858.2$^+$ & 578.2$^+$ & 637.8$^+$ & 104.4$^+$ & 80.0$^+$ & 58.6$^+$ & 168.3$^+$ & 256.0$^+$ \\
\midrule Weight (force) & 8.0 & 122.2$^+$ & 62.9$^+$ & 66.8$^+$ & 7.8 & 7.9 & 3.8 & 31.7$^+$ & 47.2$^+$ \\
Time (suddenness) & 7.8 & 146.9$^+$ & 117.6$^+$ & 95.8$^+$ & 18.3$^+$ & 8.9 & 5.6 & 21.5$^+$ & 52.1$^+$ \\
Flow (jerkiness) & 7.2 & 433.4$^+$ & 254.8$^+$ & 365.3$^+$ & 26.8$^+$ & 9.7$^+$ & 4.7 & 34.1$^+$ & 58.3$^+$ \\
Space (paths, reach) & 5.0 & 106.2$^+$ & 95.7$^+$ & 83.5$^+$ & 27.0$^+$ & 24.5$^+$ & 20.3$^+$ & 41.6$^+$ & 48.5$^+$ \\
Shape (body extent) & 1.0 & 11.2$^+$ & 6.8$^+$ & 4.3$^+$ & 6.2$^+$ & 6.2$^+$ & 4.6$^+$ & 9.1$^+$ & 8.8$^+$ \\
Body (coordination) & 4.1 & 21.6$^+$ & 24.3$^+$ & 14.8$^+$ & 14.4$^+$ & 19.2$^+$ & 13.0$^+$ & 17.1$^+$ & 28.7$^+$ \\
\bottomrule
\end{tabular}
\tablenote{Fr\'echet distance to real dance (lower is closer). Split: between two halves of the real dance's songs. 1\,s loop: real dance with one second looped. TF: Teacher Forcing; D, C: discrete-only and continuous-only. $^+$/$^-$: \method{} closer / farther than this model (95\% bootstrap CI of the difference over songs excludes zero).}
\end{table*}

FID$_{\mathrm{LMA}}$ is the Fr\'echet distance over the 55 Laban movement
descriptors of Turab et al.~\cite{turab2025dance}, and Table~\ref{tab:laban} also
gives one distance per quality: Weight (kinetic energy), Time (acceleration), Flow
(jerk), Space (paths, reach, height, travel and turning), Shape (body extent) and
Body (coordination between limbs). Windows and normalization are those of FID$_k$.
Every score separates real dance from itself with one second looped, far beyond
the distance between two halves of its songs (on the 721 sequences,
FID$_{\mathrm{LMA}}$ 1443.7 against 15.2, and 11.7--555.5 against 1.1--3.1 for
the six qualities). \cof{} lowers FID$_{\mathrm{LMA}}$
(\resultci{-69.7}{-76.1}{-63.4} against Teacher Forcing). On the 42 songs, \method{} is the closest to real dance in FID$_{\mathrm{LMA}}$ (35.0
against at least 58.6), Space, Shape and Body, with every 95\% CI excluding zero,
and on Weight, Time and Flow it does not differ significantly from MEGADance, the
closest there.

\paragraph{Baselines}
\label{sec:baseline_details}
\begin{table}[t]
\papertablestyle
\caption{Released Weights on the Same 42 Songs}
\label{tab:baselines_released}
\setlength{\tabcolsep}{1.4pt}
\begin{tabular}{@{}ll cccccc@{}}
\toprule
Method & Trained on & FID$_k\downarrow$ & Div$_k$ & Sync$\uparrow$ & MMR$\uparrow$ & BAS$\uparrow$ & Tempo$\uparrow$ \\
\midrule
Real dance & -- & -- & 6.22 & 69.0 & 0.245 & 0.247 & 80.2 \\
\midrule
\multicolumn{8}{@{}l}{\emph{Online (no future music)}} \\
\rowcolor{gray!15}\method{} & 6 datasets & 18.8 & 2.22 & \textbf{28.6} & \textbf{0.187} & 0.266 & \textbf{57.4} \\
Bailando++, strict & AIST++ & 153.5 & 14.77 & 11.9 & 0.002 & 0.250 & 56.6 \\
Danceba, strict & AIST++ & 134.8 & 16.51 & 11.9 & $-$0.006 & 0.261 & 50.9 \\
MEGADance, strict & FineDance & 79.1 & 8.44 & 11.1 & 0.003 & 0.251 & 48.2 \\
\midrule
\multicolumn{8}{@{}l}{\emph{Offline, released}} \\
EDGE & AIST++ & 24.7 & 5.97 & 16.7 & 0.010 & 0.260 & 55.2 \\
Lodge & FineDance & 21.1 & 4.32 & 14.3 & 0.008 & 0.272 & 49.6 \\
MEGADance & FineDance & \textbf{11.7} & 6.42 & 12.7 & 0.005 & 0.238 & 49.5 \\
Bailando++ & AIST++ & 148.1 & 15.66 & 14.3 & 0.001 & 0.244 & 54.8 \\
Danceba & AIST++ & 115.2 & 16.25 & 2.4 & $-$0.009 & 0.290 & 47.4 \\
\bottomrule
\end{tabular}
\tablenote{Songs, scores and \method{} as in Table~\ref{tab:baselines}, with each baseline's released weights; \method{}'s training data include other songs of this corpus. Online, strict: the released weights run on causal input, without retraining. Bold: best generator in the column; BAS is not ranked.}
\end{table}

The released generators were trained on AIST++ or FineDance. For
Table~\ref{tab:baselines}, each continues from its released weights on the
AIOZ-GDANCE training songs of \method{} (every dancer of every training
sequence; 4,808 clips, 48.3\,h) with its official training code, so it keeps
its own corpus and adds ours; none of the 42 test songs is among them. Each
stage passes over every training frame about as often as \method{}'s
training does (about 55 times), in whole epochs nearest to update targets of
3,765 for EDGE, 3,621 per stage for Lodge, 18,777 and 12,517 for MEGADance's
FSQ and GPT, 75,100 for each stage of Bailando++, and 300,424 and 75,100 for
Danceba's VQ and GPT. Token generators train at the official learning rate,
the other stages at one tenth of it. The strictly online rows use the same
weights. With the released weights (Table~\ref{tab:baselines_released}),
\method{}'s ChoreoSync offset accuracy is 28.6\% against at most 16.7\%, and
its MMR 0.187 against at most 0.010.

Bailando++ and Danceba decode greedily, as released. Consecutive motion
tokens of these training songs repeat often (26--31\%), and after training on
them the greedy decoders settle on a repeated token: 96\% of Bailando++'s
frames and 57\% of Danceba's are still (joint speed below 5\,deg/s; real dance
1\%), which lowers their Div$_k$.

Table~\ref{tab:baselines} reports no BAS. The common
implementation~\cite{siyao2022bailando} averages, over music beats, the
distance to the nearest motion beat, so a dense train of motion beats scores
high without any dance: beats every 0.2\,s and no motion score 0.848 on these
songs. Near-still motion jitters: the Bailando++ and Danceba rows of
Table~\ref{tab:baselines} have 3.2--11.1 motion beats per second against 1.0
for real dance, and BAS 0.40--0.80 against 0.247. Averaged over motion beats,
as originally defined~\cite{li2021aist}, every row and every fixed beat train
scores 0.48--0.51 on these songs.

Real dance and \method{} use the retargeting of OMG-Data; the baselines'
human motion is retargeted by a re-seeded GMR
route~\cite{araujo2025gmr}. On 39 FineDance songs with real dance through both
routes, FID$_k$ between the routes is 2.8--7.5, FID$_{\mathrm{LMA}}$ 6.4 and the
MMR difference
\resultci{-0.013}{-0.046}{0.016}; FID$_g$ between them reaches 126 and measures
the route, so Table~\ref{tab:baselines} leaves it out. ChoreoSync's estimates on
the two routes agree within 0.2\,s for 23 of 39 real dances.

The strictly online variants keep each model's start tokens and greedy
decoding. Their music features are recomputed at every generation step
from at most the last 10\,s of heard audio, with unheard frames at the model's
padding value, and their non-causal token decoders run on the 32 most recent
tokens; streaming them without look-ahead leaves seams at token boundaries
(for Bailando++, the mean absolute second difference of the pose rises from
0.0035 to 0.0305), and each token of look-ahead would add 0.133\,s of delay.
In the music-stop test, none of the strictly online variants reacts before the
change (the earliest 0.20\,s after it). Danceba outputs joint positions, which
we fit with the neutral SMPL body (median joint error about 11\,mm); both of
its variants start from the first token pair of one AIST++ training dance,
drawn per song.
MEGADance and Lodge receive three drawn styles.

We trained DiscoForcing~\cite{ji2026discoforcing} with the official code,
data and schedule and checked it against its reported FID$_k$ with the
official script before scoring any of its motion. Neither of two checkpoints
(360k steps, and a 300k-step one chosen afterwards) reached it on both test
sets within the 1.2$\times$ tolerance fixed in advance, so we compare its
interface (Table~\ref{tab:capabilities}) rather than its scores.

\paragraph{Style on unheard songs}
\label{sec:style_unheard}
\begin{table}[t]
\papertablestyle
\caption{Style on 23 FineDance Songs Unheard by \method{}}
\label{tab:style_unheard}
\setlength{\tabcolsep}{1.4pt}
\begin{tabular}{@{}llcccc@{}}
\toprule
Method & Trained on & Music & Acc. & Follow$\uparrow$ & \method{}$-$X \\
\midrule
Real dance & -- & -- & 100.0 & 47.4 & -- \\
\midrule
\multicolumn{6}{@{}l}{\emph{Online (no future music)}} \\
\rowcolor{gray!15}\method{} & 6 datasets & online & 85.5 & 32.6 & -- \\
Bailando++, strict & AIST++ & online & 39.1 & 3.0 & \resultci{29.6}{10.1}{49.0} \\
Danceba, strict & AIST++ & online & 60.9 & 6.7 & \resultci{25.9}{8.7}{45.5} \\
MEGADance, strict$^s$ & FineDance & online & 60.9 & $-$0.2 & \resultci{32.8}{12.8}{54.5} \\
\midrule
\multicolumn{6}{@{}l}{\emph{Offline, released}} \\
EDGE & AIST++ & 35\,s & 65.2 & 1.0 & \resultci{31.6}{12.8}{52.4} \\
Lodge$^s$ & FineDance & 68\,s & 73.9 & 29.6 & \resultci{3.0}{-6.9}{12.8} \\
MEGADance$^s$ & FineDance & 69.3\,s & 91.3 & 41.5 & \resultci{-8.9}{-18.8}{-1.0} \\
Bailando++ & AIST++ & 37.5\,s & 39.1 & 6.1 & \resultci{26.5}{6.7}{45.7} \\
Danceba & AIST++ & 37.5\,s & 73.9 & 19.0 & \resultci{13.6}{0.2}{29.2} \\
\bottomrule
\end{tabular}
\tablenote{FineDance songs whose audio \method{} never heard (15 Street, 8 Classic; Appendix~\ref{sec:style_unheard}). Real dance and the released generators are retargeted by the re-seeded GMR route (Appendix~\ref{sec:baseline_details}). Trained on and strict as in Table~\ref{tab:baselines}; $^s$: given each song's FineDance genre. Music: length of each song an offline generator receives; all motion is scored on its first 35.2\,s. Acc.: style accuracy (\%, descriptive). Follow: style following (points). \method{}$-$X: paired difference [95\% CI] over songs. Style following does not penalize labels absent from the song set, such as Folk here, so it can rank models differently from accuracy. Bailando++ and Danceba: one sample per song; others three. Real dance here is one dance per song from its start, on Classic and Street songs only; the 84.2\% of Table~\ref{tab:admission} covers 76 FineDance sequences of all three styles.}
\end{table}

Of the 30 FineDance songs of Fig.~\ref{fig:listenahead}(b), which \method{}
never heard, we drop the four Mix songs and the three without real dance,
leaving 23 (15 Street, 8 Classic; Table~\ref{tab:style_unheard}). \method{}
dances each from the start with three seeds; the other generators follow their
FineDance protocols, and all are scored on the first 35.2\,s. Each song is
scored by a classifier fitted without its fold or songs sharing its audio, on
six 5\,s windows from 5\,s with a majority vote. \method{}, EDGE, Lodge and offline MEGADance have every Street
sample classified as Street, so their differences come from the eight Classic
songs. On the FineDance clips of Table~\ref{tab:ablation}, which are cut from
test and validation sequences and often start mid-song, \method{}'s style
following on the 21 songs it never heard, weighted by clip, is
\resultci{5.5}{-14.6}{27.8}, not significant (real dance 46.0).

\paragraph{Whole songs}
Over whole songs the movement does not decay: on ten long FineDance songs
(108--243\,s), low activity is 7.6\% in the first minute and at most 2.8\% in
each later minute, and joint speed does not decline. In simulation SONIC
executes all ten generated songs without a fall, with a body-link position
error of 28\,mm.

\paragraph{Execution in simulation}
SONIC uses the same pretrained controller for all methods. Every simulated
trial counts, including those with falls or resets; it is scored over the
complete generated clip and succeeds if there is no reset and the pelvis-height
error stays at most 0.25\,m throughout. Position error is mean Euclidean error
over 29 pelvis-relative body links; joint error is the mean absolute joint-angle
error.

\paragraph{Music stops and returns}
\label{sec:interruption_details}
\label{sec:interruption}
EDGE, Lodge and MEGADance keep dancing through the silence at
1.03--1.77\,rad/s although they receive the whole programme in advance. Each
programme plays 20\,s of a song, interrupts it with 60\,s of silence and then
resumes it. On the 16 interruption songs from the test split, the robot
executed by SONIC in simulation responds after a median 1.03\,s, and the
continued-music control never reaches sustained low activity. Per event, the paired median
difference between measured and generated times is at most 0.07\,s in magnitude.
Across all 28 programmes the robot then holds still: it does not move again during the 60\,s silence and is at rest when the music returns. 
With the music continuing, false stops number 0 for \method{}, EDGE and the
strictly online Bailando++ and MEGADance, 2 for Lodge and the strictly online
Danceba, 6 for Bailando++, 7 for MEGADance and 11 for Danceba.

The 28 programmes cover 27 songs: the 16 test-split songs, nine from the training split and two from validation; test-split song 211 is run twice.
A response starts when the interrupted run is closer to
the standing pose than the continued run by 0.05\,rad for 0.5\,s while
moving toward it; near-standing then requires joint speed
$\le 0.3$\,rad/s and pose error $\le 0.1$\,rad for 1\,s. The standing pose
is the SONIC standing pose, which is also the generator's initial physical
boundary. Resumption is timed only for songs still at rest when the music
returns. Generated references and SONIC-measured joints are scored
on one content clock. Times exclude the
delivery buffer, set to 1.6\,s here for safety, so audio stop to measured
near-standing takes 3.92\,s on the wall clock.
With the official low-latency SONIC (0.32\,s preview) and a 0.8\,s delay,
\method{} runs the eight programmes of Appendix~\mbox{\ref{sec:physical_runs}}
in a real-time MuJoCo loop: every reference is ready at least 0.31\,s before
its deadline, no delivery deadline is missed, the pelvis stays above 0.65\,m,
and joints track with a median error of 6.3$^\circ$.

\paragraph{Physical G1}
\label{sec:physical_runs}
Twelve runs play six songs with 5\,s of silence before and after 30\,s of
music (eight runs) and two 80\,s programmes of three songs separated by 5\,s
silences (four runs), giving 20 music starts and 20 stops.
Near-standing follows each music stop after 1.47--2.00\,s, with no rebound, and
sustained dancing follows each start after 0.37--1.23\,s (medians in
Sec.~\ref{sec:execution_results}). Near-standing
uses the thresholds above on measured joints, and sustained dancing requires
joint speed above 0.3\,rad/s and a joint range over 1\,s above 0.1\,rad on
average or 0.2\,rad at the 90th percentile, for 1\,s; while the music plays, from the onset
of sustained dancing to the next stop, the robot never meets the near-standing
criterion (462\,s of dancing; at most 0.43\,s in a row). The measured delay from
reference to execution is 1.27\,s.

On a further day the music was played through the loudspeaker and captured by a
microphone beside the robot, with the same model and delivery delay. The robot
danced in eight runs, two of them on song 071, without a fall
or safety stop (supplementary video).

\paragraph{Intervals}
We use 95\% paired bootstrap intervals (10,000 draws) over songs, so that
sequences sharing a song are drawn together, averaging sampling seeds within a
sequence before resampling. Fr\'echet-distance intervals resample songs within
each dataset, recompute the set-level distance in each of 400 draws (300 for the
42 songs) and pair the two models on the same draw; Div$_k$ has no interval.
Intervals condition on the evaluated models, without multiplicity correction.

\paragraph{Timing}
\label{sec:timing_notes}
The times in Fig.~\ref{fig:robot_timeline} cover 1,200 chunks of the 12
physical runs; the slowest chunk is committed 177\,ms after its audio cutoff,
within the 0.34\,s commit deadline.

\end{document}